\documentclass[11pt]{article}

\usepackage[final]{acl}

\usepackage{times}
\usepackage{latexsym}
\usepackage{amsmath}
\usepackage{float}
\usepackage{booktabs}
\usepackage[table]{xcolor}
\usepackage{array} 
\usepackage{tabularx}
\usepackage{placeins}

\usepackage[T1]{fontenc}

\usepackage[utf8]{inputenc}

\usepackage{microtype}

\usepackage{inconsolata}

\usepackage{graphicx}

\usepackage{booktabs}
\usepackage{multirow}
\usepackage[table]{xcolor} 
\usepackage{array}

\title{HiGMem: A Hierarchical and LLM-Guided Memory System \\ for Long-Term Conversational Agents}

\author{
    \textbf{Shuqi Cao}\textsuperscript{1}\footnotemark[2],
    \textbf{Jingyi He}\textsuperscript{2}\footnotemark[2],
    \textbf{Fei Tan}\textsuperscript{1}\footnotemark[1] \\
    \textsuperscript{1}East China Normal University, Shanghai, China \\
    \textsuperscript{2}Shanghai Jiao Tong University, Shanghai, China \\
    \texttt{ftan@mail.ecnu.edu.cn}
}

\begin{document}

\maketitle
\begingroup
\renewcommand{\thefootnote}{\fnsymbol{footnote}}
\footnotetext[2]{Equal contribution.}
\footnotetext[1]{Corresponding author.}
\endgroup
\setcounter{footnote}{0}
\begin{abstract}
Long-term conversational large language model (LLM) agents require memory systems that can recover relevant evidence from historical interactions without overwhelming the answer stage with irrelevant context.
However, existing memory systems, including hierarchical ones, still often rely solely on vector similarity for retrieval. It tends to produce bloated evidence sets: adding many superficially similar dialogue turns yields little additional recall, but lowers retrieval precision, increases answer-stage context cost, and makes retrieved memories harder to inspect and manage.
To address this, we propose HiGMem (Hierarchical and LLM-Guided Memory System), a two-level event-turn memory system that allows LLMs to use event summaries as semantic anchors to predict which related turns are worth reading. This allows the model to inspect high-level event summaries first and then focus on a smaller set of potentially useful turns, providing a concise and reliable evidence set through reasoning, while avoiding the retrieval overhead that would be excessively high compared to vector retrieval.
On the LoCoMo10 benchmark, HiGMem achieves the best F1 on four of five question categories and improves adversarial F1 from 0.54 to 0.78 over A-Mem, while retrieving an order of magnitude fewer turns. Code is publicly available at \url{https://github.com/ZeroLoss-Lab/HiGMem}.
\end{abstract}

\section{Introduction}

As LLMs evolve into autonomous agents, memory systems have become essential for long-term autonomy~\cite{zhang2024survey, wang2024survey}, transforming historical data into actionable insights to maintain persona consistency and leverage past interactions for contextually-aware responses. Effective memory allows an agent to ground its responses in historical context, transforming raw interaction data into a coherent understanding of its world and relationships.

However, existing memory systems, including hierarchical ones, still
rely predominantly on vector similarity for retrieval~\cite{packer2023memgpt, xu2025mem}. While effective at casting a wide net, this approach tends to produce bloated evidence sets: once the most relevant memories have been recalled, adding more superficially similar fragments yields diminishing gains in recall but steadily erodes retrieval precision, inflates the context fed to the answer stage, and makes the evidence set harder to inspect and manage~\cite{park2023generative, lewis2020retrieval}. The root cause is that vector
similarity alone cannot judge whether a memory is truly worth reading as it lacks a mechanism for reasoning over different levels of abstraction to actively assess which fine-grained details actually contribute to answering a query. Consequently, developing retrieval strategies that can deliver concise, high-precision evidence sets remains a pivotal research objective.

To address these challenges, we introduce HiGMem, a hierarchical and LLM-guided memory system for long-term conversational agents. HiGMem tackles the core limitation of vector-only retrieval through two complementary designs. First, a two-level event-turn architecture organizes memories by abstraction level, from high-level thematic summaries to granular conversational details, providing the structural foundation for multi-granularity reasoning. Second, an LLM-guided Hierarchical Retrieval strategy leverages event summaries as semantic anchors, enabling the LLM to actively predict which fine-grained turns are worth reading rather than passively returning all similar fragments. Empirically, on the LoCoMo10 benchmark~\cite{maharana2024evaluating}, HiGMem achieves the best F1 on four of five question categories while retrieving an order of magnitude fewer turns than the strongest baseline, demonstrating that hierarchical LLM-guided retrieval can provide a concise and reliable evidence set through reasoning, while avoiding the retrieval overhead that would be excessively high compared to vector retrieval.

Our contributions are twofold: (i) We propose a retrieval paradigm for long-term conversational memory that combines hierarchical abstraction with LLM-guided filtering, enabling the system to deliver concise and reliable evidence sets through reasoning while avoiding the prohibitive cost of applying LLM reasoning to every stored memory. (ii) We implement this paradigm as HiGMem, an open-source memory system evaluated on the LoCoMo10 benchmark. HiGMem achieves the best F1 on four of five question categories while retrieving an order of magnitude fewer turns, providing a reproducible baseline for future research on long-term conversational memory.

\section{Related Work}
Long-term conversational agents depend on both strong multi-turn dialogue modeling and effective memory retrieval over long interaction histories. Recent training-time methods such as ReSURE \citep{du-etal-2025-resure} and daDPO~\citep{zhang-etal-2025-dadpo} improve multi-turn conversational ability through more reliable supervision and distribution-aware distillation. Our work focuses on the memory side of this broader problem: retrieval-time evidence localization for long-term conversational agents.

Memory-augmented LLM agents such as MemoryBank \citep{zhong2024memorybank}, MemGPT \citep{packer2023memgpt}, and A-Mem \citep{xu2025mem} extend long-horizon interaction by storing past experiences as external memories and retrieving them at inference time. However, systems that rely solely on vector-similarity retrieval tend to return broad evidence sets once the memory store grows over extended conversations. Once the most relevant evidence has been recalled, adding more semantically similar memories often brings diminishing retrieval precision, inflating context cost, and making evidence harder to inspect and manage.

Recent work has explored more structured memory organization. RAPTOR \citep{sarthi2024raptor} uses recursive summaries for multi-granularity retrieval, and concurrent work such as H-MEM \citep{sun2025hierarchical} introduces hierarchical memory routing. These studies highlight the value of structure, but prior evidence remains limited on whether long-term conversational memory systems can maintain high retrieval recall while passing only a compact set of fine-grained evidence to the answer model.

\begin{figure*}[t]
    \centering
    \includegraphics[width=\textwidth]{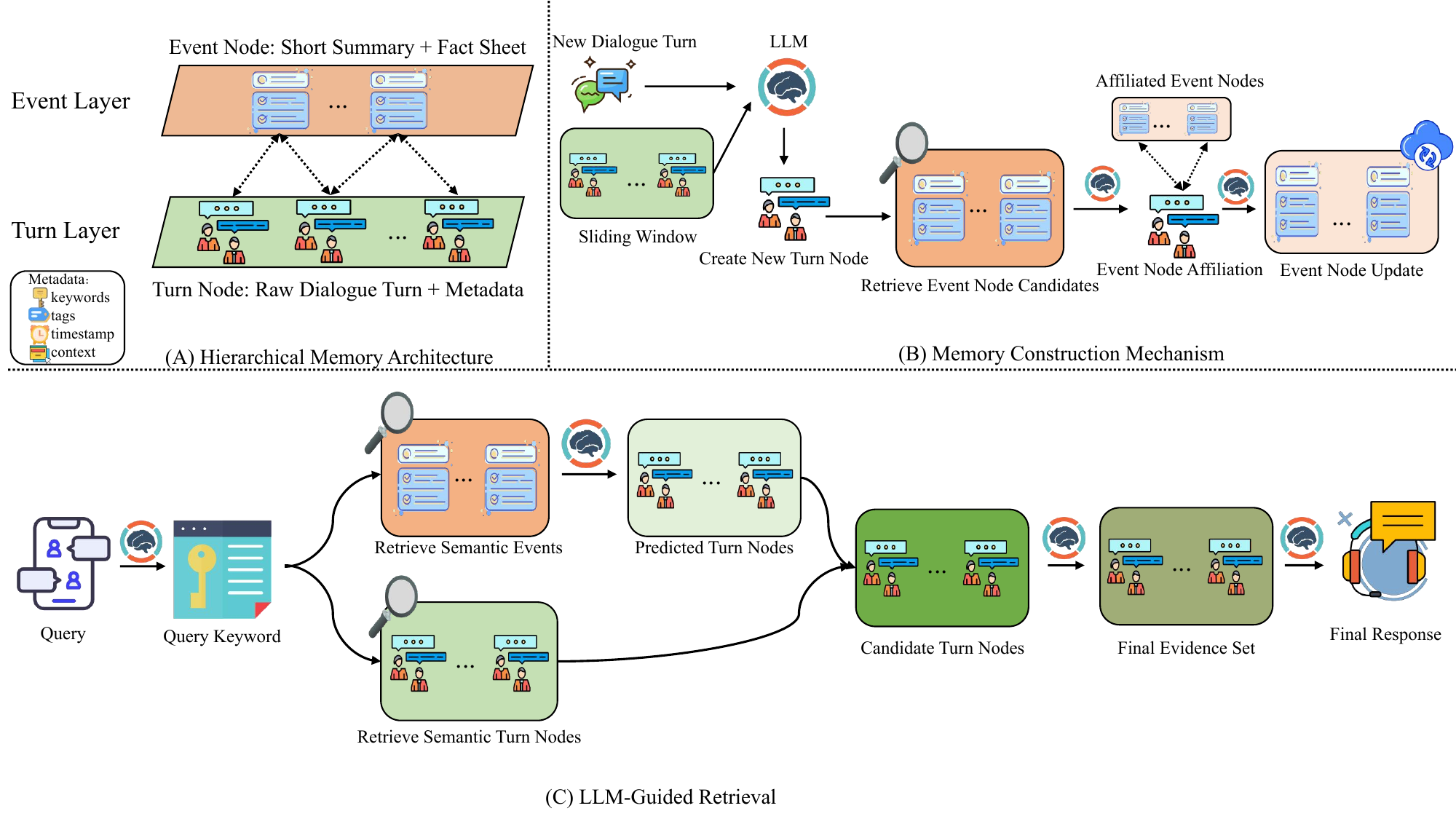} 
    
    \caption{The overall architecture of HiGMem.}
    \label{fig:main_architecture}
\end{figure*}
\section{The HiGMem System}
HiGMem is designed to address a central trade-off in long-term conversational memory retrieval: the system should recover sufficient evidence for a query, but should do so with a compact and precise evidence set rather than a large pool of semantically similar memories. We therefore pursue three objectives simultaneously: high evidence recall, high retrieval precision, and manageable token cost. To achieve this, HiGMem combines a two-level event-turn memory architecture with LLM-guided retrieval. The event layer provides concise semantic anchors that allow the LLM to inspect memory at low cost, while the turn layer preserves the fine-grained dialogue evidence required for downstream generation. Based on these event summaries, the LLM predicts which related turns are worth reading, producing a smaller and more reliable evidence set than retrieval based solely on vector similarity.

For clarity, in the remainder of this section we use \textit{Turn} node to denote HiGMem's structured memory unit built from a dialogue turn and its metadata, and \textit{Event} node to denote the higher-level unit that groups related \textit{Turn} nodes.

\subsection{Hierarchical Memory Architecture}
HiGMem features a streamlined two-level hierarchical architecture that organizes information across different granularities of abstraction. A critical component of this design is the establishment of robust bidirectional links between layers, ensuring both bidirectional interoperability and data provenance traceability.
At the foundation lies the turn layer, which captures raw interaction details at the finest granularity. Here, each \textit{Turn} node represents a raw dialogue turn enriched with LLM-generated metadata (e.g., keywords, tags, timestamp and context), serving as the primary evidence base.
Conversely, the event layer operates at the top level, grouping relevant \textit{Turn} nodes into coherent narrative units. This layer includes a short summary and a structured fact sheet. The fact sheet orchestrates the timeline and specific \textit{Event} nodes by maintaining explicit index links to the corresponding \textit{Turn} nodes in the bottom layer.

\subsection{Memory Construction Mechanism}
HiGMem supports real-time, automated memory updates. When new dialogue turn $D_t$ is ingested, the system executes an automatic update mechanism.

\paragraph{Turn Analysis.}
First, an LLM analyzes the incoming dialogue turn $D_t$ within the context of local sliding window $\mathcal{W}_t = \{T_{t-m}, \dots, T_{t-1}\}$ ($m$ denotes the sliding window size) to extract metadata. Metadata $\mathcal{M}_t$ includes keywords, tags, timestamp and context. After analyzing, new \textit{Turn} node $T_t$ consists of dialogue turn $D_t$ and metadata $\mathcal{M}_t$. This transformation can be formalized as:

\begin{equation}
    D_t \xrightarrow{\text{LLM}(\cdot \mid \mathcal{W}_t)} \mathcal{M}_t,
    \label{eq:turn_processing}
\end{equation}

\begin{equation}
    T_t = (D_t, \mathcal{M}_t),
    \label{eq:turn_definition}
\end{equation}

\noindent where the arrow in Eq.~\eqref{eq:turn_processing} denotes the generative transformation by the LLM.

\paragraph{Event Affiliation and Update.}
Simultaneously, the system determines the affiliation of $T_t$ at the event layer. 
We employ a text encoder to compute the dense vector representation of $T_t$, denoted as $e_t$. 
Subsequently, we calculate similarity score $s_{t,j}$ between $T_t$ and each existing \textit{Event} node $E_j$ in the global set of \textit{Event} nodes $\mathcal{E}$ using cosine similarity:

\begin{equation}
\label{eq:sim_formula}
s_{t,j} = \frac{e_t \cdot e_{E_j}}{\|e_t\| \|e_{E_j}\|}, \quad \forall E_j \in \mathcal{E},
\end{equation}
where $e_{E_j}$ represents embedding vector of \textit{Event} node $E_j$. 
We retrieve the top-$k_{\text{event}}$ most relevant events to construct candidate set $\mathcal{E}_{\text{cand}}$.
The set of \textit{Event} nodes to which $T_t$ is affiliated $\mathbf{E}^*$ is identified by an LLM-based decision function:

\begin{equation}
    \mathbf{E}^* = f_{\text{affiliate}}(T_t \mid \mathcal{E}_{\text{cand}}),
\end{equation}

\noindent where $\mathbf{E}^* \subseteq \mathcal{E}_{\text{cand}}$ represents a subset of \textit{Event} nodes. 
Subsequently, for each \textit{Event} node $E \in \mathbf{E}^*$, we perform an adaptive update based on the scale of the interaction. 
Specifically, based on \textit{Event} node volume $|E|$ (i.e., the number of linked \textit{Turn} nodes in \textit{Event} node $E$) and threshold $\tau$, we fully refresh both the summary and fact sheet if $|E| < \tau$, but only append the new entry if $|E| \ge \tau$.
Finally, to ensure traceability, we update link set $\mathcal{L}_E$ for each affiliated \textit{Event} node by appending new link $l$ between $E$ and $T_t$:

\begin{equation}
\mathcal{L}_E = \mathcal{L}_E \cup \{l\}, \quad \forall E \in \mathbf{E}^*.
\end{equation}

In all experiments, sliding window size for turn analysis is fixed to $m=5$, and adaptive \textit{Event} node update threshold is fixed to $\tau=10$.

\subsection{LLM-Guided Retrieval}
To answer query $Q$, HiGMem employs an LLM-guided retrieval strategy.
The system first analyzes $Q$ to generate query keyword $q_{\text{kw}}$: $q_{\text{kw}} = \text{LLM}(Q)$.
Based on Eq.~\eqref{eq:sim_formula}, query $q_{\text{kw}}$ simultaneously retrieves the $k_{\text{turn}}$ most relevant \textit{Turn} nodes ($\mathcal{T}_{\text{semantic}}$) and the $k_{\text{event}}$ most relevant \textit{Event} nodes ($\mathcal{E}_{\text{semantic}}$) from the turn and event layers, respectively.
Uniquely, HiGMem utilizes $\mathcal{E}_{\text{semantic}}$ as semantic anchors for further discovery.
The LLM evaluates constituent \textit{Turn} nodes $\mathcal{T}_{\text{E}}$ which are linked to each \textit{Event} node $E$ in $\mathcal{E}_{\text{semantic}}$, performing reasoning to select specific \textit{Turn} nodes capable of answering the query. This process yields inference-based set $\mathcal{T}_{\text{pred}}$:

\begin{equation}
    \mathcal{T}_{\text{pred}} = \bigcup_{E \in \mathcal{E}_{\text{semantic}}} \text{Predict}(\mathcal{T}_E \mid Q,E).
\end{equation}
The system integrates $\mathcal{T}_{\text{semantic}}$ and $\mathcal{T}_{\text{pred}}$, and then employs an LLM-based filtering mechanism to construct final evidence set, denoted by $\mathcal{T}_{\text{final}}$, which consists of \textit{Turn} nodes:

\begin{equation}
    \mathcal{T}_{\text{cand}} = \mathcal{T}_{\text{semantic}} \cup \mathcal{T}_{\text{pred}},
\end{equation}
    
\begin{equation}
    \mathcal{T}_{\text{final}} = \text{Filter}( \mathcal{T}_{\text{cand}}\mid Q),
\end{equation}
where $\text{Filter}(\cdot \mid Q)$ denotes the filtering process.
Finally, \textit{Turn} nodes in the final evidence set $\mathcal{T}_{\text{final}}$ and query $Q$ are fed into the LLM to generate a response.


\begin{table*}[t]
\centering
\footnotesize 
\renewcommand{\arraystretch}{0.95} 
\setlength{\tabcolsep}{3pt} 

\newcolumntype{C}{>{\centering\arraybackslash}X}

\begin{tabularx}{0.92\textwidth}{l CCCCCC}
\toprule
\textbf{Method} & \textbf{Multi-Hop} & \textbf{Temporal} & \textbf{Open-Domain} & \textbf{Single-Hop} & \textbf{Adversarial} & \textbf{Ranking} \\
\midrule
Base LLM & 0.25 & \textbf{0.39} & 0.12 & 0.44 & 0.30 & 2.2 \\
ReadAgent & 0.09 & 0.13 & 0.05 & 0.10 & 0.10 & 4.2 \\
MemoryBank & 0.05 & 0.10 & 0.06 & 0.07 & 0.07 & 4.8 \\
A-Mem & 0.27 & \textbf{0.39} & 0.10 & 0.42 & 0.54 & 2.2 \\
\rowcolor{gray!15}
\textbf{HiGMem (Ours)} & \textbf{0.31} & 0.34 & \textbf{0.15} & \textbf{0.49} & \textbf{0.78} & \textbf{1.2} \\
\bottomrule
\end{tabularx}
\caption{Experimental results on the LoCoMo10 dataset across five question categories, reported in F1 score. Ranking is the average category-wise rank across the five question categories, where lower is better. The best performance is highlighted in bold.}
\label{tab:main_results}
\end{table*}

\section{Experiments}
\subsection{Experimental Setup}
\noindent\textbf{Dataset}. We evaluate HiGMem on LoCoMo10, a subset of 10 longest conversations with high-quality annotations selected by the authors of LoCoMo \citep{maharana2024evaluating} from the full LoCoMo dataset, with an average of 587 turns per conversation. LoCoMo10 is particularly suitable for our setting because it contains significantly longer multi-session dialogues than existing conversational datasets~\citep{xu-etal-2022-beyond,jang-etal-2023-conversation}, requiring models to recover evidence over extended interaction histories. Based on its question answering (QA) benchmark, we evaluate memory retrieval across five categories: Single-Hop, Multi-Hop, Temporal, Open-Domain, and Adversarial questions.

\noindent\textbf{Baselines}. We compare HiGMem against ReadAgent \citep{lee2024human}, MemoryBank \citep{zhong2024memorybank}, A-Mem \citep{xu2025mem}, and a full-context GPT-4o-mini baseline following prior LoCoMo10 evaluations. We do not include H-MEM \citep{sun2025hierarchical} in the main quantitative comparison because we could not identify an official public implementation, and key answer-stage details are insufficient for fair reproduction.

\noindent\textbf{Metrics}. We report F1 score \citep{rajpurkar2016squad} and retrieval metrics, including Avg K, Precision@K, and Recall@K. Here, $K$ denotes the number of turns in the final evidence set provided to the answer-generation LLM for each question. Precision@K is the fraction of retrieved turns that match gold evidence turns, while Recall@K is the fraction of gold evidence turns recovered by the final evidence set. Since a question may require one or more gold evidence turns, we compute Precision@K and Recall@K at the question level and report their macro-average over questions, rather than pooling evidence turns across all questions.

\noindent\textbf{Implementation}. For a controlled comparison, all LLM-based systems use GPT-4o-mini \citep{achiam2023gpt}, and all methods use the same final QA prompt for answer generation. HiGMem and A-Mem use the same all-MiniLM-L6-v2 \citep{reimers2019sentence} encoder for vector retrieval. For HiGMem, we set $k_{\text{turn}}=10$ and $k_{\text{event}}=10$ for turn- and event-level retrieval. We summarize the main prompt families in Appendix~\ref{sec:appendix_prompts}; complete prompt templates, implementation constants, and evaluation scripts are provided in the released code.

\subsection{Results and Analysis}
Table \ref{tab:main_results} summarizes performance on the LoCoMo10 benchmark~\citep{maharana2024evaluating}. HiGMem achieves the best F1 on four of five categories, including Multi-Hop, Open-Domain, Single-Hop, and Adversarial questions. Its Temporal F1 is slightly lower than A-Mem, suggesting that the current event-level abstraction may still weaken some fine-grained chronological cues. Overall, the results show that HiGMem is effective in supporting question answering over long conversational histories.

\begin{table}[t]
\centering
\small
\begin{tabular}{lccc}
\toprule
\textbf{Method} & \textbf{Avg K} & \textbf{Precision@K} & \textbf{Recall@K} \\
\midrule
A-Mem & 99.84 & 0.0101 & \textbf{0.7502} \\
\rowcolor{gray!15}
\textbf{HiGMem} & \textbf{8.09} & \textbf{0.1909} & 0.7241 \\
\bottomrule
\end{tabular}
\caption{Retrieval effectiveness on LoCoMo10. Avg K denotes the average number of turns in the final evidence set provided to the answering-generation LLM. HiGMem achieves comparable recall with a much smaller and more precise evidence set.}
\label{tab:retrieval_summary}
\end{table}

To analyze the source of the F1 score gains, Table \ref{tab:retrieval_summary} directly evaluates the quality of the retrieved evidence set. Compared with A-Mem, HiGMem retrieves an order of magnitude fewer turns on average (8.09 vs. 99.84), while achieving much higher Precision@K (0.1909 vs. 0.0101) and comparable Recall@K (0.7241 vs. 0.7502). This supports our central claim: HiGMem does not simply increase context size to improve recall, but instead produces a compact and more precise evidence set.

\begin{table}[b]
\centering
\small
\begin{tabular}{lcc}
\toprule
\textbf{Method} & \textbf{A-Mem} & \textbf{HiGMem} \\
\midrule
GPT-4o-mini Tokens & \textbf{11.81M} & 54.35M \\
GPT-5 Answer Tokens & 25.38M & \textbf{1.62M} \\
GPT-4o-mini Cost & \textbf{\$1.44} & \$5.30 \\
GPT-5 Cost & \$15.99 & \textbf{\$1.12} \\
Total Cost & \$17.43 & \textbf{\$6.43} \\
\bottomrule
\end{tabular}
\caption{End-to-end cost analysis under a hybrid deployment setting, where memory construction and retrieval use GPT-4o-mini, while final answer generation uses GPT-5. HiGMem consumes more tokens before answer generation, but substantially reduces the expensive GPT-5 answer-stage input.}
\label{tab:end_to_end_cost}
\end{table}

Table \ref{tab:end_to_end_cost} further clarifies the token cost trade-off. HiGMem uses more GPT-4o-mini tokens during memory construction and retrieval, but substantially reduces the expensive GPT-5 answer-stage input from 25.38M to 1.62M tokens in a hybrid deployment setting. As a result, the estimated total cost decreases from \$17.43 to \$6.43, about a 2.7$\times$ reduction. Table~\ref{tab:time_cost} reports the average time overhead.
HiGMem requires roughly 2.4$\times$ more time for the memory
construction stage and 1.6$\times$ more for the question answering stage due to
additional LLM reasoning calls, a trade-off consistent with
the token-cost shift discussed above.

\begin{table}[t]
\centering
\small
\begin{tabular}{lcc}
\toprule
\textbf{Time Metric} & \textbf{A-Mem} & \textbf{HiGMem (Ours)} \\
\midrule
Memory Construction (s) & \textbf{6.38} & 15.59 \\
Question Answering (s) & \textbf{5.91} & 9.42 \\
\bottomrule
\end{tabular}
\caption{Comparison of computational efficiency in terms of average per-instance time overhead. Memory construction is measured per ingested dialogue turn, while question answering is measured per question.}
\label{tab:time_cost}
\end{table}

Additional controlled experiments are reported in Appendix~\ref{sec:appendix_b}, including a flat retrieval baseline with LLM selection, fixed-budget retrieval comparisons with A-Mem, a supplementary DialSim generalization study, and a Qwen2.5-3B backbone comparison.

\subsection{Ablation Study}
To validate the role of the event-turn hierarchy, we remove the event layer and degrade HiGMem to a single-layer memory system. As shown in Table~\ref{tab:ablation_study}, removing the hierarchy reduces both F1 and Recall@K. This indicates that the event layer is not merely a storage abstraction: event summaries provide compact semantic anchors that reduce the turn-level search space and help the LLM select more relevant fine-grained evidence.

\begin{table}[t]  
\centering
\small
\setlength{\tabcolsep}{3.5mm}
\begin{tabular}{lcc}
\toprule
\textbf{Method} & \textbf{F1} & \textbf{Recall@K} \\
\midrule
w/o Hierarchy & 0.39 & 0.55 \\ 
\rowcolor{gray!15} 
\textbf{HiGMem} & \textbf{0.49} & \textbf{0.72} \\
\bottomrule
\end{tabular}
\caption{
Ablation study to verify the effectiveness of the hierarchical structure. 
}
\label{tab:ablation_study}
\end{table}

\section{Conclusion}
We present HiGMem, a hierarchical memory system for long-term conversational agents that addresses a central retrieval trade-off: recovering sufficient evidence while keeping the evidence set compact and precise. HiGMem combines an event-turn memory architecture with LLM-guided retrieval, using event summaries as semantic anchors to predict which fine-grained turns are worth reading. On LoCoMo10, HiGMem achieves the best F1 on four of five question categories while retrieving far fewer turns than A-Mem, and substantially reduces answer-stage cost in hybrid deployment. These results suggest that LLM-guided evidence selection is an effective direction for building long-term conversational memory systems.

\section*{Limitations}
HiGMem can reduce downstream answer-stage cost by producing a compact final evidence set, but it still requires additional LLM calls for memory construction and LLM-guided evidence selection. Therefore, its practical advantage depends on the deployment setting: it is most beneficial when a lightweight model handles memory operations while a more expensive downstream model performs final answer generation. Reducing the time and token overhead of these intermediate LLM calls while preserving evidence precision remains an important direction.

HiGMem also depends on the LLM's ability to judge relevance from event summaries and candidate turns. Although supplementary experiments with Qwen2.5-3B show a consistent trend on LoCoMo10, the preliminary DialSim study shows lower F1 than A-Mem despite a much smaller evidence set, suggesting that the current event-turn structure is not yet fully optimized for multi-party dialogue. Broader evaluation on real-world, multi-party, noisy, and dynamically evolving conversations is needed to fully assess robustness.


\section*{Acknowledgments}
This work is supported by the East China Normal University ``Artificial Intelligence'' Seed Grant Program (40500-20101-222438) and the East China Normal University ``Discipline Advancement Program'' (40600-515100-25001/002/015).

\bibliography{main}
\clearpage
\appendix

\section{Illustrative Comparison of Retrieval Paradigms}
\label{app:paradigm_comparison}


\begin{figure}[htb] 
\centering
\small
\begin{tabular}{>{\raggedright\arraybackslash}p{0.95\linewidth}}
\toprule
\textbf{Query:} "What kind of car does Evan drive?" \\[1ex] 
\midrule

\textbf{Paradigm 1: Passive Vector Retrieval} \\
\textit{(e.g., RAG, A-Mem)} \\[0.5ex]
1. \textbf{Reduce:} Query is reduced to keywords: \{"kind", "car", "Evan", "drive"\}. Intent is lost. \\[0.5ex]
2. \textbf{Search:} A single vector search retrieves semantically similar but factually irrelevant memories (e.g., "a two-hour drive"). \\[0.5ex]
3. \textbf{Result:} \textcolor{red}{Failure}. The system cannot find the specific "Prius" evidence. \\
\midrule

\rowcolor{gray!15}
\textbf{Paradigm 2: LLM-Guided Retrieval (HiGMem)} \\
\rowcolor{gray!15}
\rowcolor{gray!15}
1. \textbf{Reason:} LLM analyzes the full query, identifying retrieval goals: find a \textit{thing} (`car`) related to an \textit{entity} (`Evan`). \\[0.5ex]
\rowcolor{gray!15}
2. \textbf{Predict:} The LLM evaluates the query against event summaries to identify which linked turns potentially serve as the answer. \\[0.5ex]
\rowcolor{gray!15}
3. \textbf{Locate:} Based on these predictions, the system pinpoints the exact evidence within the candidate turns.\\[0.5ex]
\rowcolor{gray!15}
4. \textbf{Result:} \textcolor{blue}{Success}. The system retrieves turns mentioning "my new Prius". \\
\bottomrule
\end{tabular}
\caption{
A comparison of retrieval paradigms. 
Passive Vector Retrieval suffers from semantic drift due to keyword reduction. 
In contrast, HiGMem's LLM-guided Retrieval leverages reasoning to predict relevant turns from event summaries and locate exact evidence, ensuring accurate recall.
}
\label{fig:paradigm_comparison_appendix}
\end{figure}

\section{Supplementary Experiments} 
\label{sec:appendix_b}
\subsection{Flat Retrieval + LLM Selection Baseline}
\label{sec:appendix_b1}

We remove the event layer entirely: flat vector search retrieves the top-100 turns, then applies the same LLM relevance filtering and final answer prompt as HiGMem. This controlled baseline tests whether the gain comes merely from LLM filtering or from using event summaries as semantic anchors.

\begin{table}[h]
\centering
\small
\begin{tabular}{lcc}
\toprule
\textbf{Task / Metric} & \textbf{Flat+LLM} & \textbf{HiGMem} \\
\midrule
Multi-Hop & 0.29 & \textbf{0.31} \\
Temporal & \textbf{0.37} & 0.34 \\
Open-Dom & 0.11 & \textbf{0.15} \\
Single-Hop & 0.46 & \textbf{0.49} \\
Adversarial & 0.73 & \textbf{0.78} \\
\midrule
\textbf{Overall F1} & 0.46 & \textbf{0.49} \\
\bottomrule
\end{tabular}
\caption{Comparison of task-specific and overall F1 scores between the Flat Retrieval + LLM Selection baseline and HiGMem.}
\label{tab:app_exp1_f1_vertical}
\end{table}

\begin{table}[h]
\centering
\small
\begin{tabular}{lcc}
\toprule
\textbf{Metric} & \textbf{Flat+LLM} & \textbf{HiGMem} \\
\midrule
Avg K & 22.7 & \textbf{8.09} \\
Precision@K & 0.065 & \textbf{0.190} \\
Recall@K & 0.703 & \textbf{0.724} \\
\bottomrule
\end{tabular}
\caption{Retrieval efficiency and prompt cost comparison.}
\label{tab:app_exp1_retrieval_vertical}
\end{table}

Compared with HiGMem, the Flat+LLM baseline retrieves a substantially larger evidence set (22.71 vs. 8.09 turns on average), but still achieves lower overall F1 (0.46 vs. 0.49), lower Precision@K (0.065 vs. 0.190), and slightly lower Recall@K (0.703 vs. 0.724). This shows that the event layer is not merely an implementation detail for enabling LLM filtering: Event summaries act as semantic anchors that reduce the turn-level search space and improve evidence precision.

\subsection{Comparison with A-Mem at Fixed K}
\label{sec:appendix_b2}

We truncate A-Mem's retrieved turns (including neighbor expansion) at fixed K, and compare with HiGMem's natural output.

\begin{table}[h]
\centering
\small
\begin{tabular}{lccc}
\toprule
\textbf{Method} & \textbf{K} & \textbf{Precision@K} & \textbf{Recall@K} \\
\midrule
A-Mem & 8 & 0.059 & 0.385 \\
A-Mem & 16 & 0.038 & 0.478 \\
A-Mem & 32 & 0.024 & 0.580 \\
A-Mem & 99.84 (full) & 0.010 & 0.750 \\
\midrule
HiGMem & 8.09 (natural) & \textbf{0.190} & \textbf{0.724} \\
\bottomrule
\end{tabular}
\caption{Fixed-K retrieval comparison with A-Mem. A-Mem requires a much larger expanded context to approach HiGMem's natural-output recall.}
\label{tab:app_exp2_amem}
\end{table}

\subsection{End-to-End Cost Analysis}
\label{sec:appendix_b3}

To substantiate the claim that retrieving fewer turns effectively reduces downstream inference costs, we conduct an end-to-end cost analysis on the LoCoMo10 dataset. In the following analysis, ``avg K'' refers to the average number of turns ultimately provided to the answer-generation LLM per question, and ``MC'' denotes Memory Construction.

\begin{table}[htb]
\centering
\small

\begin{tabular}{lcc}
\toprule
\textbf{Metric} & \textbf{A-Mem} & \textbf{HiGMem} \\
\midrule
Overall F1 & 0.37 & \textbf{0.49} \\
Avg K (Answer Stage) & 99.84 & \textbf{8.09} \\
\midrule
MC Tokens & \textbf{9.2M} & 21.2M \\
MC Time & \textbf{12.8h} & 23.6h \\
QA Tokens & \textbf{25.5M} & 29.3M \\
QA Time & \textbf{2.4h} & 4.2h \\
\bottomrule
\end{tabular}
\caption{Breakdown of F1 score, retrieval size, token usage, and time consumption during the memory construction (MC) and question answering (QA) stages.}
\label{tab:app_exp3_breakdown}

\vspace{1.5em} 

\begin{tabular}{lcc}
\toprule
\textbf{Cost Component} & \textbf{A-Mem} & \textbf{HiGMem} \\
\midrule
Mini Tokens (MC+Ret.) & \textbf{11.8M} & 54.5M \\
GPT-5 Tokens (Answer) & 25.4M & \textbf{1.6M} \\
\midrule
Mini Cost & \textbf{\$1.44} & \$5.30 \\
GPT-5 Cost & \$15.99 & \textbf{\$1.12} \\
\midrule
\textbf{Total Cost} & \$17.43 & \textbf{\$6.43} \\
\bottomrule
\end{tabular}
\caption{Deployment cost simulation (MC and retrieval utilizing GPT-4o-mini, while answer generation utilizes GPT-5).}
\label{tab:app_exp3_cost}

\end{table}

\paragraph{Token Shift.} As detailed in Table \ref{tab:app_exp3_breakdown}, HiGMem spends more tokens and time during memory construction and retrieval-stage reasoning than A-Mem. However, under the hybrid deployment accounting in Table \ref{tab:app_exp3_cost}, this cost is shifted away from the expensive answer model: GPT-5 answer tokens fall from 25.386M to 1.629M, i.e., from roughly 12.8K to 0.8K tokens per question.

\paragraph{Deployment Cost Savings.} Table \ref{tab:app_exp3_cost} simulates a deployment in which memory construction and retrieval use GPT-4o-mini, while final answer generation uses GPT-5. Because HiGMem passes a much smaller final evidence set to the expensive answer model, the estimated total cost drops from \$17.43 to \$6.43, about a 2.7$\times$ reduction.

\subsection{Generalization to DialSim Dataset}
\label{sec:appendix_b4}

To evaluate the generalization capability of our approach to a second dataset, we conduct preliminary experiments on DialSim v1.1, a multi-party TV show dialogue dataset (including \textit{Friends}, \textit{The Big Bang Theory}, and \textit{The Office}). We sampled 7,000 turns and 3,000 questions across three shows under a shared manifest. Note that HiGMem was not specifically tuned for this multi-party setting.

\begin{table}[htb]
\centering
\small

\begin{tabular}{lcc}
\toprule
\textbf{Metric} & \textbf{A-Mem} & \textbf{HiGMem} \\
\midrule
Overall F1 & \textbf{0.49} & 0.42 \\
Avg K & 95.8 & \textbf{3.9} \\
\midrule
MC Prompt Tokens & \textbf{9.3M} & 25.2M \\
MC Time & \textbf{9.1h} & 15.1h \\
QA Prompt Tokens & 36.7M & \textbf{23.5M} \\
QA Time & \textbf{2.9h} & 4.0h \\
\bottomrule
\end{tabular}
\caption{F1 score, retrieval size, and resource breakdown on the DialSim dataset (3,000 questions).}
\label{tab:app_exp4_breakdown}

\vspace{1.5em}

\begin{tabular}{lcc}
\toprule
\textbf{Cost Component} & \textbf{A-Mem} & \textbf{HiGMem} \\
\midrule
Mini Tokens (MC+Ret.) & \textbf{12.0M} & 52.5M \\
GPT-5 Tokens (Answer) & 36.5M & \textbf{1.4M} \\
\midrule
Mini Cost & \textbf{\$1.45} & \$5.10 \\
GPT-5 Cost & \$22.96 & \textbf{\$1.06} \\
\midrule
\textbf{Total Cost} & \$24.42 & \textbf{\$6.17} \\
\bottomrule
\end{tabular}
\caption{Deployment cost on DialSim (GPT-4o-mini for MC/Retrieval, GPT-5 for Answer).}
\label{tab:app_exp4_cost}

\end{table}

As shown in Table \ref{tab:app_exp4_breakdown}, A-Mem achieves a higher F1 score (0.49 vs. 0.42), which is primarily due to its much larger retrieval context (avg K=95.8 vs. 3.9). However, this does not diminish our core contribution of LLM-driven reasoning-based retrieval and memory structuring. The highly compact average $K$ demonstrates that our retrieval paradigm remains effective at precise evidence localization. The F1 gap may reflect that the current two-level event-turn structure is not yet fully optimized for multi-party dialogue. Despite the lower F1, Table \ref{tab:app_exp4_cost} shows that HiGMem's downstream token efficiency advantage remains robust, proving to be $\sim$4.0$\times$ cheaper overall in a hybrid deployment scenario. Exploring more generalizable memory structures for multi-party settings is a promising direction for future work.

\subsection{Storage and Retrieval Scaling}
\label{sec:appendix_b5}

To address potential concerns regarding storage and retrieval efficiency as the memory size grows, we analyze the scaling behaviors of both methods. Note that the retrieval time here measures pure vector-only computation (cosine top-K) and excludes LLM API calls.

\begin{table}[h]
\centering
\small
\begin{tabular}{lrr}
\toprule
\textbf{turns} & \textbf{A-Mem} & \textbf{HiGMem} \\
& (Storage / Latency) & (Storage / Latency) \\
\midrule
100  & 0.20MB / 0.038ms & 0.26MB / 0.042ms \\
10K  & 15.4MB / 0.15ms  & 16.7MB / 0.20ms \\
100K & 153.6MB / 4.5ms  & 166.7MB / 4.8ms \\
1M   & 1536MB / 47.7ms  & 1667MB / 53.0ms \\
\bottomrule
\end{tabular}
\caption{Scaling analysis of storage footprint and vector-only retrieval latency across different memory sizes.}
\label{tab:app_exp5_scaling}
\end{table}

 As illustrated in Table \ref{tab:app_exp5_scaling}, both approaches scale linearly with the number of turns. HiGMem incurs approximately 8.5\% additional storage overhead due to the required event-layer data. Furthermore, the vector-only retrieval latency remains well within a $\sim$10\% margin compared to A-Mem at all scales. This demonstrates that the hierarchical structure of HiGMem introduces negligible computational overhead to the underlying vector search process.

\subsection{Open-Source Backbone: Qwen2.5-3B}
We further evaluate HiGMem with Qwen2.5-3B on \textit{LoCoMo10}, comparing against A-Mem under the same backbone. This experiment was run without query rewriting, so it is intended as a supplementary robustness check rather than the main reported setting. The trend remains consistent: HiGMem improves overall F1 while retrieving substantially fewer turns.

\begin{table}[h]
\centering
\small
\begin{tabular}{lcc}
\toprule
\textbf{Metric} & \textbf{A-Mem} & \textbf{HiGMem} \\
\midrule
Overall F1 & 0.34 & \textbf{0.42} \\
Avg K & 76.54 & \textbf{21.68} \\
Precision@K & 0.011 & \textbf{0.051} \\
Recall@K & 0.634 & \textbf{0.657} \\
\bottomrule
\end{tabular}
\caption{Supplementary Qwen2.5-3B comparison on \textit{LoCoMo10}.}
\label{tab:app_qwen25_3b}
\end{table}

\section{Prompt Summaries}
\label{sec:appendix_prompts}

We provide the complete prompt templates in the released code. In this appendix, we summarize the four prompt families most important for reproducing HiGMem on LoCoMo10: turn analysis, event affiliation, retrieval-stage filtering, and category-specific final QA generation. The summaries below describe the practical role, input structure, and output format of each prompt family, rather than reproducing the full templates verbatim.

\paragraph{Turn Analysis Prompt.}
This prompt is used when a new dialogue turn is written into memory. Its input consists of a short sliding window of recent dialogue turns together with the current turn, including speaker, timestamp, and text. The model is asked to analyze the current turn semantically. The output is a structured JSON object containing metadata (e.g., keywords, context, timestamp and tags), profile-retrieval keys, and direct links. This prompt supports the turn layer by converting raw dialogue into a structured memory unit that can later participate in retrieval and event construction.

\paragraph{Event Affiliation Prompt.}
This prompt determines whether a newly added turn should attach to one or more existing events or start a new event. Its input includes the new turn and a small set of top candidate events retrieved by similarity, together with optional recent event context. The model is asked to compare the new turn against these candidate events and output a structured affiliation decision, including selected event IDs or \texttt{NEW\_EVENT}. This prompt is the main mechanism that builds the event layer online and groups related turns into higher-level memories rather than leaving the memory as a flat set of turns.

\paragraph{Retrieval-Stage Filtering Prompt.}
This prompt family is used during question answering to reduce a broad candidate pool into the final evidence turns shown to the answer-generation model. Its input includes the user question and candidate memory items gathered from vector retrieval, event-based prediction, and optional link expansion. The model is asked to judge which candidate items are useful for answering the question, with special attention to subject matching, topical relevance, and temporal constraints. The output is a JSON object containing the IDs of candidate items judged relevant. Before this final filtering step, HiGMem also performs an event-local selection step that asks the model which turns inside a retrieved event are most likely to contain the answer. Together, these prompts turn a large candidate pool into a smaller and more precise evidence set.

\paragraph{Final QA Generation Prompt.}
For the LoCoMo10 setting used in our main HiGMem runs, final QA generation uses a concise category-specific prompt family rather than a single generic prompt. All variants take as input the retrieved turn context and the question; adversarial questions additionally include the distractor answer candidate. For temporal questions, the model is instructed to infer an approximate date from the dialogue timeline and answer briefly. For adversarial questions, the model is constrained to choose between the candidate answer and \textit{Not mentioned in the conversation}. For the remaining categories, the model is instructed to answer with a short phrase and to use exact words from the retrieved context whenever possible. These prompts ensure short, category-appropriate, and evidence-grounded final answers.

\end{document}